\documentclass[runningheads]{llncs}
\usepackage{xcolor}
\usepackage{amssymb}
\usepackage{amsmath} 
\usepackage{comment}
\usepackage{pgfplots}
\usepackage{booktabs}
\usepackage{algorithm}
\usepackage{algorithmic}
\usepackage{arydshln}
\usepackage{multirow}

\usepackage{graphicx}
\newcommand{\antonio}[1]{\textcolor{blue}{{\bf [Antonio: }{\em #1}{\bf ]}}}
\newcommand{\yashar}[1]{\textcolor{green!55!blue}{{\bf [Yashar: }{\em #1}{\bf ]}}}

\newcommand{\tommaso}[1]{\textcolor{magenta}{{\bf [Tommaso: }{\em #1}{\bf ]}}}

\definecolor{ultramarine}{RGB}{0,32,96}
\definecolor{neworange}{RGB}{255,87,51}

\newcommand{\dquotes}[1]{``#1''}

\begin{document}
\title{Towards Effective Device-Aware\\Federated Learning}
%\title{Towards Effective Personalized Federated Learning using a Priority-Aware Aggregation Operator and Dynamic Parameter Updating}
%
\titlerunning{Towards Effective Device-Aware Federated Learning}
% If the paper title is too long for the running head, you can set
% an abbreviated paper title here
%
\author{
Vito Walter Anelli \and Yashar Deldjoo \and
Tommaso Di Noia \and Antonio Ferrara\thanks{Corresponding author.}}
\authorrunning{ }
% First names are abbreviated in the running head.
% If there are more than two authors, 'et al.' is used.
%
\institute{Polytechnic University of Bari, Bari, Italy \\
\email{firstname.lastname@poliba.it}}
\maketitle              % typeset the header of the contribution
%
%%%% Proceedings format for most of ACM conferences (with the exceptions listed below) and all ICPS volumes.
%\documentclass[sigconf]{acmart}

%
% The abstract is a short summary of the work to be presented in the article.
\begin{abstract}
%Over the last few years, we have witnessed a remarkable breakthrough in capabilities of artificial intelligence (AI) algorithms, approaching or exceeding human-level performance across a note-worthy number of tasks. Much of this success owes to massive amount of data empowering deep neural network models. Nevertheless, 
With the wealth of information produced by social networks, smartphones, medical or financial applications, speculations have been raised about the sensitivity of such data in terms of users' personal privacy and data security.
To address the above issues, Federated Learning (FL) has been recently proposed as a means to leave data and computational resources distributed over a large number of nodes (clients) where a central coordinating server aggregates 
only locally computed updates without knowing the original data. In this  work, we extend the FL framework by pushing forward the state the art in the field on several dimensions: (i) unlike the original FedAvg approach relying solely on single criteria (i.e., local dataset size), a suite of \textit{domain-} and \textit{client-specific criteria} constitute the basis to compute each local client's contribution, (ii) the multi-criteria contribution of each device is computed in a prioritized fashion by leveraging a \textit{priority-aware aggregation operator} used in the field of information retrieval, and (iii) a mechanism is proposed for \textit{online-adjustment} of the aggregation operator parameters via a local search strategy with backtracking. Extensive experiments on a publicly available dataset indicate the merits of the proposed approach compared to standard FedAvg baseline.

\end{abstract}

\section{Introduction and Context}
\label{sec:intro}
The vast amount of data generated by billions of mobile and online IoT devices worldwide holds the promise of significantly improved usability and user experience in intelligent applications. This large-scale quantity of rich data has created an opportunity to greatly advance the intelligence of machine learning models by catering powerful deep neural network models. Despite this opportunity, nowadays such pervasive devices can capture a lot of data about the user, information such as what she does, what she sees and even where she goes~\cite{DBLP:journals/itpro/MillerVH12}. Actually, most of these data contain sensitive information that a user may deem private.
%Recently concerns have been expressed about sensitivity of this \yashar{chg to user?} data in terms of user's privacy and data security. 
To respond to concerns about sensitivity of user data in terms of data privacy and security, in the last few years, initiatives have been made by governments to prioritize and improve the security and privacy of user data. For instance, in 2018, General Data Protection Regulation (GDPR) was enforced by the European Union to protect users' personal privacy and data security. These issues and regulations pose a new challenge to traditional AI models where one party is  involved in collecting, processing and transferring all data to other parties. As a matter of fact, it is easy to foresee the risks and responsibilities involved in storing/processing such sensitive data in the traditional \textit{centralized} AI fashion.

Federated learning is an approach recently proposed by Google~\cite{konevcny2015federated,konevcny2016federated,mcmahan2017communication} with the goal to train a global machine learning model from a massive amount of data, which is \textit{distributed} on the client devices such as personal mobile phones and/or IoT devices. 
%FL shares the idea of the general concept \dquotes{bringing the code to the data, instead of data to code} in which 
In principle, a FL model is able to deal with fundamental issues related to privacy, ownership and locality of data~\cite{DBLP:journals/corr/abs-1902-01046}.  In~\cite{mcmahan2017communication}, authors introduced the \textit{FederatedAveraging} (FedAvg) algorithm, which combines local stochastic gradient descent on each client via a central server that performs model aggregation by averaging the values of local hyperparameters. To ensure that the developments made in FL scenarios uphold to real-world assumptions, in~\cite{caldas2018leaf} the authors introduced LEAF, a modular benchmarking framework supplying developers/researchers with a rich number of resources including open-source federated datasets, an evaluation framework, and a number of reference implementations.

%\yashar{@Tommaso: All the following parts have been substantially revised/improved. It needs one round of careful check by you.}

Despite its potentially disruptive contribution, we argue that FedAvg exposes some major shortcomings. First, the aggregation operation in FedAvg sets the contribution of each agent proportional to each individual client's local dataset size. A wealth of qualitative measures such as the number of sample classes held by each agent, the divergence of each computed local model from the global model --- which may be critical for convergence~\cite{sahu2018convergence} ---, some estimations about the agent computing and connection capabilities or about their honesty and trustworthiness are ignored. While FedAvg only uses limited knowledge about local data, %\yashar{maybe we can say: leading to imprecise aggregation?} \antonio{I would not take these extreme sides...}\yashar{sure}, 
we argue that the integration of the above-mentioned qualitative measures and the expert's domain knowledge is indispensable for increasing the quality of the global model. 

%by only considering the information about 
%local data and 
%and disregards a wealth of information about models, agents as well as local data available which we assume could be exploited to improve the performance of the global model.
%by asking the agents for few information \yashar{remove the last part I think we already mentioned it in First}\tommaso{OK}. %\yashar{@Anotnio: this part seems to be correlated with the last statement.} Finally, FedAvg treats the federation process as statistic learning problem and does not allow a mechanism for adjustment of the aggregation operator parameters. \antonio{I don't know if this statement is necessary... especially because it can be a little bit confusing about aggregation} 
%\yashar{I revised the last paragraph to connect it better to our work. The last sentence can be improved.}
%\yashar{@Tommaso: The following is newly added}\\
The work at hand  considerably extends the FedAvg approach~\cite{mcmahan2017communication} by building on three main assumptions:

\begin{itemize}
    \item we can substantially improve the quality of the global model by incorporating \textit{a set of criteria} about domain and clients, and properly assigning the contribution of individual update in the final model based on these criteria;

    %First: we hypothesize that integration a set of \textit{carefully chosen criteria} about domain and clients and exploiting them for assignment of each individual's contribution to the final model can lead to (substantial) improvements in terms of effectiveness and efficiency of the final global model;
    
    %proper assignment of the contribution of each individual local update in the final model, we can substantially improve the quality of the global model;
    
     %\antonio{what about this?}\yashar{good choice} 
    \item the introduced criteria can be combined by using different aggregation operators; toward this goal, we assert about the potential benefits of using a \textit{prioritized multi-criteria aggregation operator} over the identified set of criteria to define each individual's local update contribution to the federation process;
    
    %going beyond averaging; %\yashar{Would be illustrative if we could mention an example, in the literature for example they give and example between safety and cost. Safety always has a higher priority compared with cost. An example related to our criteria could be useful if possible.}
    %which can \yashar{be exploited to} substantially impact the final result \yashar{please be more specific in terms of what or we can say quality of the global model}. 
 
    %based \antonio{both on the criteria fulfillment and on the priority of each criterion};
    
    %assess/understand the degree of satisfaction of a prioritized \yashar{Do we need to say prioritized again?} set of criteria for defining each individual's local update contribution to the federation process; \yashar{Please check the second contribution} 

    %the use of an aggregation operator for appropriately gathering the information about each criterion measurement;
    
    %\item we hypothesize that a key to better exploitation of criteria relies on \textit{how their information is exploited for individual contribution assignment} for which we consider the effects of a more effective aggregation procedure important factor; \yashar{We can even improve this sentence by saying: we hypothesize existence of a prioritization relationship over the identified set of criteria ...}
    
    \item computation of parameters for the aggregation operator (the priority order of the above-mentioned criteria) via an \textit{online monitoring and adjustment} is an important factor for improving the quality of global model. %\yashar{The semantic of this sentence is not clear, maybe we want to say "the aggregation operator parameters REQUIRE an \textit{online monitoring and adjustment} and, which we deem to play an important role in enhancing the quality of global model?}
\end{itemize}
The remainder of the paper is structured as follows. Section~\ref{sec:FL} is devoted to introducing the proposed FL system, it first describes the standard FL model and then provides a formal description of the proposed FL approach and the key concepts behind integration of local criteria and prioritized multi-criteria aggregation operator in the proposed system. Section~\ref{sec:expe} details the  experimental setup of the entire system by relying on LEAF, an open-source benchmarking framework for federated settings, which comes with a suite of datasets realistically pre-processed for FL scenarios. Section~\ref{subsec:eval_FL_proposed} presents results and discussion. Finally, Section~\ref{sec:conlusion} concludes the paper and discusses future perspectives.

\section{Federated Learning and Aggregation Operator}
\label{sec:FL}
%\yashar{Our goal here is to introduce formalism to the work via defining FL problem and showing what the problem is and where our contribution resides. In section 3.2, we define where aggregation operator is injected.}
In the following, we introduce the main elements behind the proposed approach. We start by presenting a formal description to the standard FL approach (cf. Section~\ref{sec:FL_BKG}) and then we describe our proposed FL approach (cf. Section~\ref{sec:FL_proposed}). 
\subsection{Background: Standard FL}
\label{sec:FL_BKG}
In a FL setup, a set $\mathcal{A}=\{A_1, ..., A_K\}$ of agents (clients) participate to the training federation with a server $S$ coordinating them. Each agent $A_k$ stores its local data $\mathcal{D}_k = \{(x_1^k,y_1^k), (x_2^k,y_2^k), ..., (x_{|\mathcal{D}_k|}^k,y_{|\mathcal{D}_k|}^k) \}$, and never shares them with $S$. In our setting, $x_i^k$ represents the data sample $i$ of agent $k$ and $y^k_i$ is the corresponding label. The motivation behind a FL setup is mainly efficiency --- $K$ can be very large --- and privacy~\cite{bagdasaryan2018backdoor,mcmahan2017communication}. As local training data $\mathcal{D}_k$ never leaves federating agent machines, FL models can be trained on user private (and sensitive) data, e.g., the history of her typed messages, which can be considerably different from publicly accessible datasets.

 The final objective in FL is to learn a global model characterized by a 
 %\antonio{global (i would remove this repetition)}\yashar{thank you} 
 parameter vector $\mathbf{w}^G \in \mathbb{R}^{d}$, with $d$ being the number of parameters for the model, such that a global loss is minimized without a direct access to data across clients. The basic idea is to train the global model separately for each agent $k$ on $\mathcal{D}_k$, such that a local loss is minimized and the agents have to share with $S$ only the computed model parameters $\mathbf{w}^k$, which will be aggregated at the server level. 
 %\antonio{Here I would avoid to mention the variable name, since the formalism (also taking into account the time dimension) is presented below} \yashar{YES I just meant the bold style, now it is ok.}
 
 %The basic idea is to train a local model for agent $k$ on $\mathcal{D}_k$ such that it minimizes a local loss and the only information exchanged with $S$ is the computed local parameters $\mathbf{w}^k$. \yashar{Please check the last paragraph if it read fluent.} \antonio{I would change the last sentence with something like this: "The basic idea is to train the global model separately for each agent $k$ on $\mathcal{D}_k$, such that a local loss is minimized and the agents have to share with $S$ only the computed parameters $w^k$ which will be aggregated at the server level."}\yashar{very good}
 
 %The basic idea is, to train a local model characterized by local parameter $\mathbf{w}^k$ for agent $k$ on $\mathcal{D}_k$ such that it minimizes a local loss, and exchange with $S$ only information about the computed parameters $\mathbf{w}^k$. 

By means of a communication protocol, the agents and the global server exchange information about the parameters of the local and global model.
At the $t$-th round of communication, the central server $S$ broadcasts the current global model $\mathbf{w}^G_t$ to a fraction of agents $\mathcal{A}^- \subset \mathcal{A}$. 
%\tommaso{Are we sure? To my understanding, the server broadcasts to all the agents in $\mathcal{A}$ but only the fraction in $\mathcal{A}^-$ returns their updates to the server. Otherwise there are agents in $\mathcal{A} $ that never receive updates.} \antonio{Actually, in literature this is the way the original idea is described, at least for describing the training process (you can have a look also at~\cite{konevcny2016federated2}, which we can also cite in the article, as they neither take into account the dataset size for aggregation)... probably in a realistic implementation it could be appropriate to send the global results to all the devices (but also in this case there are many factors, like the online status of the devices)..}. 
Then, every agent $k$ in $\mathcal{A}^-$ carries out some optimization steps over its local data $\mathcal{D}_k$ in order to optimize a local loss. Finally, the computed local parameter vector $\mathbf{w}^k_{t+1}$ is sent back to the central server. The central server $S$ computes a weighted mean of the resulting local models in order to obtain an updated global model $\mathbf{w}^G_{t+1}$
\begin{equation}
    \label{eq:SGD}
    \mathbf{w}^G_{t+1} = \sum_{k=1}^{|\mathcal{A}^-|} p^k_{t+1} \mathbf{w}^k_{t+1}.
\end{equation}
%\yashar{Maybe here we can mention that clients can get the updated version of the global model (e.g., during update) to improve their local models.}\\
For the sake of simplicity of discussion, throughout this work, we do not consider the time dimension and focus our attention on one time instance as given by Equation (\ref{eq:SGD_simple})

\begin{equation}
    \label{eq:SGD_simple}
    \mathbf{w}^G = \sum_{k=1}^{|\mathcal{A}^-|} p^k \mathbf{w}^k,
\end{equation}
in which $p^k \in [0,1]$ is the weight associated with agent $k$ and $\sum_{k=1}^{|\mathcal{A}^-|} p^k = 1$.

We argue that collecting information about clients and incorporating that knowledge to compute the appropriate agent-dependent value $p^k$ is important for computing an effective and efficient federated model. Moreover, it is worth noticing that $p^k$ may encode and carry out some useful knowledge in the optimization of the global model with respect to relevant domain-specific dimensions.

\subsection{Proposed Federated Learning Approach}
\label{sec:FL_proposed}
As discussed at the end of the previous section, we may have different factors and/or criteria influencing the computation of $p^k$. Given a set of properly identified criteria about clients, it could be then possible to enhance the global model update procedure by using this information.

%\antonio{in order to establish the importance of each device --- i.e. its contribution to the federated model --- as if it were a multi-criteria decision making problem.}

%\antonio{I reformulated this part}
%A stated in Section~\ref{sec:intro}, one of the main novelties of the present work is
%the determination of the contribution of each agent w.r.t. the new global model as if it were a multi-criteria decision making (MCDM) problem~\cite{Triantaphyllou2000,GRABISCH1996445} with respect to the identified set of criteria.

%The contribution of each agent is estimated by incorporating the information about a set of domain-, data- and agent-depending criteria. \yashar{i think first we should mention criteria and then aggregation.} 

To connect it to the formalism presented before, let us assume $C=\{C_1,...,C_m\}$ be a set of measurable properties (criteria) characterizing local agent $k$ or local data $\mathcal{D}_k$.  We use the term $c_i^k \in [0,1]$ to denote, for each agent $k$, the degree of satisfaction of criterion $C_i$ in a specific round of communication. Hence, in the proposed FL aggregation protocol, the central server computes $p^k$ as 
\begin{equation}
\label{eq:score}
    p^k = \frac{f(c_1^k,...,c_m^k)}{Z} = \frac{s^k}{Z},
\end{equation}
%\yashar{XXX To improve readability, I added the normalization term equal to $Z$. The variable can be named anyway.}\antonio{ok!}\\
\noindent where $f$ is a \textit{local aggregation operation}  over the set of properties (criteria), which represent agent $k$, $s^k \in \mathbb{R}$ is a numerical score evaluating the $k$-th agent contribution based on the $m$ identified properties and, finally, $Z$ is a normalization factor. In order to ensure that $\sum_{k=1}^{|\mathcal{A}^-|}p^k=1$ where $p^k \in [0,1]$, we compute $Z=\sum_{k=1}^{|\mathcal{A}^-|} s^k$.

In the following, we briefly discuss the identified set of criteria (together with a motivation for the selection), the selected aggregation operator, and the online adjustment procedure. 

\bigskip
\noindent\textbf{Identification of local criteria.}
%Our approach makes it possible to use whatever criterion considered useful to the training of the global model as well as to the efficiency of the learning procedure. 
%\yashar{The original sentences here seemed a bit informal. I improved and suggested the following, please check:}\\
In FedAvg, the server performs aggregation to compute $p^k$, without knowing any information about participating clients, except for a pure quantitative measure about local dataset size. 
%\antonio{what do you think now?}\yashar{It is ok to me.}.
Our approach relies on the assumption that it might be much better to use multiple criteria encoding different useful knowledge about clients to obtain a more informative global model during training. This makes it possible for a domain expert to build the federated model by leveraging different any additional \textit{domain-} and \textit{client-specific} knowledge. 
%As we already mentioned, our approach provides a domain expert with the capability to build the federated model by leveraging different \textit{domain-} and \textit{client-specific} knowledge. 

For instance, one may want to choose the criteria in such a way that the rounds of communication needed to reach a desired target accuracy are minimized. %\yashar{@Tommaso: is it possible for you check the following part. It could connect with personalization/fairness and it is good to have an overall check. } 
Moreover, a domain expert could ask users/clients to measure their adherence to some other target properties (e.g. their nationality, gender, age, job, behavioral characteristics, etc.), in order to build a global model emphasizing the contribution of some classes of users; in this way, the domain expert may, in principle,  build a model favoring some targeted commercial purposes.

All in all, we may have a suite of criteria to reach the final global goal (in Section \ref{subsec:iden_criteria} we will see the example adopted in our experimental setup).
%as we will see in our experimental setup presented in Section \ref{subsec:iden_criteria}.
%we show our  experimental evaluation 
%In the following we present 
%This paper integrates a suite of criteria which we will present as part of our experimental setup in Section .

%In our experimental part, we investigated with a very small suite of criteria, just aiming to maximize the percentage of devices receiving in less rounds of communication a model which performs well with their own test data. Details about this experimentation are provided in Section \ref{sec:expe}.
\bigskip
\noindent\textbf{Prioritized multi-criteria aggregation operator.}
%existence of a prioritization relationship over the identified set of criteria
%\yashar{I think the following we can reduce.}
%\antonio{Please, check the following... I introduced some other AO} 
Once local criteria evaluations have been collected, the central server aggregates  them for each device in order to obtain a final score associated to that device. 
%\yashar{Should we say that the server is responsible for this? Once the local criteria about domain and clients are identified, the server has to aggregate them in order to compute the contribution of each agent.}
Over the years, a wide range of aggregation operators have been proposed in the field of information retrieval (IR)~\cite{pasi}. We selected some prominent ones and exploited them in our FL setup. In particular, we focused on  the weighted averaging operator, the ordered weighted averaging (OWA) models~\cite{DBLP:journals/ijis/Yager96,DBLP:journals/tsmc/Yager88}, which extend the binary logic of \textit{AND} and \textit{OR} operators by allowing representation of intermediate quantifiers, the Choquet-based models~\cite{AIF195451310,Grabisch2000,GRABISCH1996445}, which are able to interpret positive and negative interactions between criteria, and finally the priority-based models~\cite{dacosta}. 
Due to the lack of space, here we report only the approach and the experimental evaluation related to the last one, modeled in terms of a MCDM problem, because of its better performance. 

%we will consider as our choice in the remainder of this paper.

The core idea of the \textit{prioritized multi-criteria aggregation operator} proposed in~\cite{dacosta} is to assign a priority order to the involved criteria. The main rationale behind the idea is to allow a domain expert to model circumstances where the lack of fulfillment of a higher priority criterion cannot be compensated with the fulfillment of a lower priority one~\cite{pasi}. As an example, we may consider the case where the domain expert may want to consider extremely important the age of an agent's user rather than its dataset size, so that even a large local dataset would be penalized if the user age criteria is not satisfied.
%\yashar{till here: for example we can simply say there exists different aggregation operators like simple statistics based on average or more sophisticated ones based xxx , xxx, xxx. In this work, we chose xxx. The fundamental assumption of xxx is existence of a prioritization relationship over the identified set of criteria. The example we can shorten.}

Formally, the prioritized multi-criteria aggregation operator $f: [0,1]^m \rightarrow [0,m]$ measures an overall \textit{score} from a prioritized set of criteria evaluations on the local model $\mathbf{w}^k$ as in the following~\cite{dacosta}:

%is presented for deriving an overall \textit{score} from a prioritized set of  criterion evaluations on the local model $\mathbf{w}^k$ \yashar{given by}

%In~\cite{dacosta}, a prioritized multicriteria aggregation operator $f_1: [0,1]^m \rightarrow [0,m]$ is presented for deriving an overall \textit{score} from a prioritized set of  criterion evaluations on the local model $\mathbf{w}^k$ \yashar{given by}

\begin{equation}
\label{eq:pasi}
\begin{split}
    s^k = f(c_1^k,...,c_m^k) = \sum_{i=1}^m \lambda_i \cdot c_{(i)}^k \\
    \lambda_1 = 1, \quad  \lambda_i = \lambda_{i-1}\cdot c_{(i-1)}^k, \enspace i \in [2,m]
    \end{split}
\end{equation}

\noindent where $c_{(i)}^k$ is the evaluation of $C_{(i)}$ for device $k$ and the $\cdot_{(i)}$ notation indicates the indices of a sorted priority order for criteria, as specified by the domain expert, from the most important to the least important one. For each score $c_{(i)}^k$, an importance weight $\lambda_i$ is computed, depending both on the specified priority order over the criteria and on the fulfillment and the weight of the immediately preceding criterion. 
\begin{example}
Let us suppose that we are interested in evaluating device $k$ based on three criteria $C_1,C_2,C_3$ and their respective evaluations are $c_1^k = 0.5, c_2^k = 0.8, c_3^k = 0.9$. Let the priority order of criteria be $C_{(1)}=C_1, C_{(2)}=C_2, C_{(3)}=C_3$, from the most important to the least important; then, $\lambda_1 = 1, \lambda_2 = \lambda_1\cdot c_{(1)}^k=0.5, \lambda_3 = \lambda_2\cdot c_{(2)}^k = 0.4$. Hence, the final device score will be $s^k = (1 \cdot 0.5) + (0.5 \cdot 0.8) + (0.4 \cdot 0.9) = 1.26$. If we change the priority order to be $C_{(1)}=C_3, C_{(2)}=C_2, C_{(3)}=C_1$, we would then obtain  $\lambda_1 = 1, \lambda_2 = \lambda_1\cdot c_{(1)}^k=0.9, \lambda_3 = \lambda_2\cdot c_{(2)}^k = 0.72$ with a final device score of $s^k = (1 \cdot 0.9) + (0.9 \cdot 0.8) + (0.4 \cdot 0.5) = 1.82$. We see that this latter value is higher than the previous one since the most important criterion here is better fulfilled. \hfill\(\Box\)
%\antonio{Pls, have a look at the example I put here}
\end{example}

%Furthermore, the weight computed for criterion $i$ (with a higher priority) depends on both the weights and xxx \yashar{@anotio: please complete }. 

\noindent\textbf{Online adjustment.}
%\yashar{XXX SHALL WE MENTION OUR GOAL SIMPLY HERE?  XXX WE CAN SIMPLY THE INTRODUCTION ABOVE A BIT AND SAY SIMPLY: OUR GOAL IS TO COMPUTE THE BEST CRITERIA PRIORITY ORDER FOR THE AGGREGATION OPERATOR XXX FOR ALL AGENTS PARTICIPATING TO XXX. THIS IS COMPUTED AT EACH ROUND OF COMM BY SERVER (STH ALONG THIS)}
%Our approach provides a formal protocol for making the most of criteria evaluations we computed in the previous steps. 
%\antonio{Please, check it now} 
%\yashar{this part I have not read yet, I should read it.}
The aggregation operator we are using takes as parameter the priority order of the involved criteria and, as a consequence, one of the problem is to identify the best ordering for Equation~\ref{eq:pasi} which takes benefit of the gathered information. Although by definition this priority order could be defined by a domain expert, here we propose to choose the best one in an online fashion such that we can  maximize the performances of the model at each round of communication.

%} For this reason, we propose a method which computes at each round of communication the priority order maximizing the performances of the model which, in turn, is then adopted for each device participating to that round of communication. 
%In the remainder of this paragraph, we present our protocol at the $t$-th round of communication.

Let $(C_{(1),t},...,C_{(m),t})$ be the last priority ordering of the criteria used to compute the local scores $p^k_t$ (see Equation~(\ref{eq:score})~and~(\ref{eq:pasi})) at time $t$. 
%for building $\mathbf{w}^G_t$. 
The sequence of steps needed to compute the updates to the global model is formalized in Algorithm~\ref{alg:algoritmo} and commented in the following. 

\begin{algorithm}[t!]
\caption{Sequence of steps executed by the server to compute the new global model with online adjustment of aggregation operator parameters. Functions \textit{ModelUpdate}, \textit{PropertyMeasure}, and \textit{LocalTestAccuracy} are executed locally on the $k$-th device. Variable $\textrm{acc}_t$ is an estimation of the global accuracy.}
\begin{algorithmic}[1]
\label{alg:algoritmo}
 \REQUIRE $\mathbf{w}^G_t$, $\textrm{acc}_t$, $(C_{(1),t},...,C_{(m),t})$
 \ENSURE $\mathbf{w}^G_{t+1}$, $\textrm{acc}_{t+1}$, $(C_{(1),t+1},...,C_{(m),t+1})$ 
 %\STATE \textbf{Server executes:}
\STATE broadcast $\mathbf{w}^G_t$ to clients in $\mathcal{A}^-$
 \FOR {each client $k \in \mathcal{A}^-$ \textbf{in parallel}}
 \STATE $\mathbf{w}^k_{t+1} \leftarrow \textrm{ModelUpdate(} k, \mathbf{w}^G_t \textrm{)}$
 \FOR {each criterion $C_i \in C$}
 \STATE $c_{i,t+1}^k \leftarrow \textrm{PropertyMeasure(} k, \mathbf{w}^k_{t+1}, C_i \textrm{)}$
 \ENDFOR
  \ENDFOR
  \STATE $P \leftarrow (C_{(1),t},...,C_{(m),t})$
 \FOR {each client $k \in \mathcal{A}^-$}
 \STATE $p^k_{t+1} \leftarrow  f(c_{(1),t+1}^k,...,c_{(m),t+1}^k) /Z$
   \ENDFOR
 \STATE $\mathbf{\overline{w}}^G_{t+1} \leftarrow \sum_{k=1}^{|\mathcal{A}^-|} p^k_{t+1} \mathbf{w}^k_{t+1}$
  \FOR {each client $k \in \mathcal{A}$ \textbf{in parallel}}
 \STATE $\textrm{acc}^k_{t+1} \leftarrow \textrm{LocalTestAccuracy(} k, \mathbf{\overline{w}}^G_{t+1} \textrm{)}$
\ENDFOR
%\STATE  $\textrm{acc}_{t+1} \leftarrow \sum_{k=1}^{|\mathcal{A}|} \textrm{acc}^k_{t+1}$
\STATE $\textrm{acc}_{t+1} \leftarrow$ weighted average of $\textrm{acc}^k_{t+1}$ w.r.t. local test set size, $\forall k \in \mathcal{A}$
\WHILE {$\textrm{acc}_{t+1} < \textrm{acc}_t$}
\IF {other priority orderings are available}
  \STATE $P \leftarrow$ another priority ordering of criteria $(C_{(1)},...,C_{(m)})\star$
\STATE repeat steps 9---16
\ELSE
\STATE $P \leftarrow$ priority ordering for which we get the maximum value for $\textrm{acc}_{t+1}$ 
\STATE $\textrm{acc}^k_{t+1} \leftarrow$ accuracy of the model which performed best
\STATE repeat steps 9---12
\STATE \textbf{break}
\ENDIF
\ENDWHILE
  \STATE $(C_{(1),t+1},...,C_{(m),t+1}) \leftarrow P$
  \STATE  $\mathbf{w}^G_{t+1} \leftarrow \mathbf{\overline{w}}^G_{t+1} $
\end{algorithmic}
\end{algorithm}

\begin{description}
    \item[Lines 1--7] On each device, we locally train the last broadcasted global model $\mathbf{w}^G_t$ with the local training data, in order to compute $\mathbf{w}^k_{t+1}$; then, we measure the local scores for each of the identified criteria.
    \item[Lines 9--11] For each device, we use the priority ordering of criteria already used in the previous round of communication to compute the local score $p^k_{t+1}$.
    \item[Line 12] A new \textit{candidate} global model $\mathbf{\overline{w}}^G_{t+1}$ is built by computing a weighted averaging of the local models w.r.t. the computed $p^k_{t+1}$.
    \item[Lines 13--15] On each device, $\mathbf{\overline{w}}^G_{t+1}$ is locally tested using the local test set.
    \item[Lines 16--29] An estimation of a global accuracy is computed weighting local accuracies w.r.t. local test set size; then, if the obtained accuracy is higher on average than the accuracy obtained with $\mathbf{w}^G_t$, then we update the global value $\mathbf{w}^G_{t+1} \leftarrow \mathbf{\overline{w}}^G_{t+1}$ and we proceed with the next round of communication; otherwise, another permutation is considered and, once a new $p^k_{t+1}$ is computed for each device, we go back to step 3; if no other permutations are available, the candidate global model which produced the least worst test accuracy is assigned to $\mathbf{w}^G_{t+1}$.
\end{description}

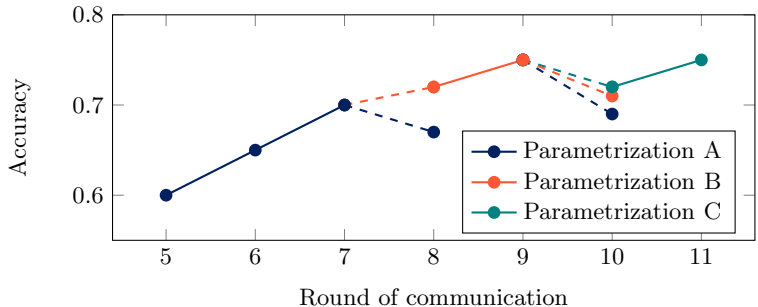
\begin{figure}[tb]
\begin{center}
%\resizebox{0.9\textwidth}{!}{%
\begin{tikzpicture}
\begin{axis} [xlabel={Round of communication}, ylabel={Accuracy}, xtick={5,6,7,8,9,10,11}, ymin=0.55, ymax=0.8, scale only axis, % The height and width argument only apply to the actual axis
width=0.7\textwidth,height=3cm, legend pos=south east]

\addplot[ultramarine, thick, mark=*]
coordinates  {(5,0.60) (6,0.65) (7,0.7)};
\addlegendentry{Parametrization A}
\addplot[neworange, thick, dashed, mark=*, mark options={solid},forget plot]
coordinates  {(7,0.7) (8,0.72)};
\addplot[ultramarine, thick, dashed, mark=*, mark options={solid},forget plot]
coordinates  {(7,0.7) (8,0.67)};
\addplot[neworange, thick, mark=*]
coordinates  {(8,0.72) (9,0.75)};
\addlegendentry{Parametrization B}
\addplot[teal, thick, dashed, mark=*, mark options={solid},forget plot]
coordinates  {(9,0.75) (10,0.72)};
\addplot[ultramarine, thick, dashed, mark=*, mark options={solid},forget plot]
coordinates  {(9,0.75) (10,0.69)};
\addplot[neworange, thick, dashed, mark=*, mark options={solid},forget plot]
coordinates  {(9,0.75) (10,0.71)};
\addplot[teal, thick, mark=*]
coordinates  {(10,0.72) (11,0.75)};
\addlegendentry{Parametrization C}

\end{axis}
\end{tikzpicture}
%}
\end{center}
\caption{An illustration of the online parameter adjustment for the  aggregation operator. }\label{fig:plot-online}
\end{figure}

 The above-mentioned steps are also graphically illustrated by means of a plot in Figure~\ref{fig:plot-online}, where an exemplification with dummy values is presented. Training steps proceed with the same parametrization until a lower  accuracy 
 %\yashar{***important: I am not sure if we should mention test accuracy VIVIDLY here, in practice we do not have a test accuracy and we are not allowed to use. Maybe we can say validation accuracy or some how leave it a bit fuzzy? In future, we can create a validation set from train data and perform these experiments} \antonio{Actually, we use the local test sets to measure accuracy at each round of comm.} 
 is obtained (blue point in round of communication 8); then, the previous model is restored and the other configurations are tested, until a higher  accuracy is found (e.g., orange point in round 8). When a higher accuracy cannot be found, the least worst option is selected (e.g., green point in round 10).

\section{Experimental setup}
\label{sec:expe}

In this section we describe the experimental setup used to validate the  performance of the proposed FL system. 

\bigskip
%First, we introduce LEAF~\cite{caldas2018leaf} xxxx. Next, ... \yashar{t.b.c.}  

\noindent\textbf{Experimental Evaluation Framework.}
\label{exp_appr}
In order to perform the experimental validation and performance evaluation, an extensive set of experiments has been carried out by relying on LEAF~\cite{caldas2018leaf}, a modular open-source benchmarking framework for federated settings, which comes with a suite of datasets appropriately preprocessed for FL scenarios. LEAF also provides reproducible reference implementations and introduces both system and statistical rigorous metrics for understanding the quality of the FL approach.

As for the metrics computation, the global model is tested on each device over the local test sets. The objective of LEAF is to capture the distribution of performance across devices by considering the 10th and 90th percentiles of the local accuracy values and by estimating a global accuracy (local accuracy values are averaged weighting them based on local test set size).

In this work, we improve the validation of the FL setting by using an approach which offers an overview of the whole training performances, instead of metrics describing a single round of communication. More specifically, \textit{we measure the number of round of communication required to allow a certain percentage of devices, which participate to the federation process, to reach a target accuracy (e.g., 75\% or 80\%)}, since this measurement is able to fairly show how effective and efficient is the model across the devices.

%\yashar{nice, could we make a practical example? We deem this to be important for deployment in practical settings, where xxx (we can make an example)}

% LEAF also recognizes the importance of specifying how the accuracy is weighted across devices, e.g., whether every device is equally important, or every data point equally important (implying that power users/devices get preferential treatment). Notably, considering stratified systems and accuracy metrics is particularly important in order to evaluate whether a method will systematically exclude groups of users (e.g., because they have lower end devices) and/or will underperform for segments of the population (e.g., because they produce less data).
\bigskip
\noindent\textbf{Federated dataset.}
%Instead of using the trivial MNIST dataset~\cite{MNIST} also used in~\cite{mcmahan2017communication},
We run our experiments using the FEMNIST dataset~\cite{caldas2018leaf}, which contains  handwritten characters and digits from various writers and their true labels. Unlike the original FedAvg algorithm~\cite{mcmahan2017communication}, which uses the MNIST dataset~\cite{MNIST} artificially split by labels, the FEMNIST dataset~\cite{caldas2018leaf}, is larger and more realistically distributed. The dataset contains 805,263 examples of 62 classes of handwritten characters and digits from 3,550 writers and it is built by partitioning data in ExtendedMNIST~\cite{cohen2017emnist} --- an extended version of MNIST with letters and digits --- based on writers of digits/characters. It is important to note that data in FEMNIST are inherently non-IID distributed, as the local training data can vary between clients; therefore, they are not representative of the whole population distribution. We use the described dataset to perform a digit/character classification task, although for computational limits we use a subsampled version (10\% of total, 371 clients involved).

\bigskip
\noindent\textbf{Convolutional model.}
Similar to~\cite{mcmahan2017communication}, the classification task is performed by using a convolutional neural network (CNN). The network has two convolutional layers with 5x5 filters  --- the first with 32 channels, the second with 64, each followed by 2x2 max pooling ---, a fully connected layer with 2048 units and ReLu activation, and a final softmax output layer, with a total of 6,603,710 parameters.

\bigskip
\noindent\textbf{Hyperparameter settings.} 
We set the hyperparameters for the whole set of our experiments as follows, also guided by the results obtained in~\cite{mcmahan2017communication}. As for the FedAvg client fraction parameter, in each round of communication only 10\% of clients are selected to perform the computation. For what concerns the parameters of stochastic gradient decent (SGD), we set the local batch size to 10 and the number of local epochs equal to 5. This is the configuration that in the baseline makes it possible to reach the target accuracy in less rounds of communication. Moreover, we set the learning rate to $\eta = 0.01$. Finally, we set the maximum number of rounds of communication per each experiment to 1000.

\bigskip
\noindent\textbf{Identified local criteria.}
\label{subsec:iden_criteria}
In our experimental setting, the proposed FL system extends pure quantitative criteria in FedAvg~\cite{mcmahan2017communication}  --- dataset size ---  and leverages two new criteria. Please note that we are not stating that the proposed ones are the only possible criteria. We present them just to show how the introduction of new information may lead to a better final model. More specifically, in our experimental evaluation, we aim at both reducing the number of rounds of communication necessary to reach a target accuracy and making the global model not diverging towards local specializations and overfittings.

The criteria have been defined so that $c_i^k \in [0,1]$ with $0$ meaning bad performance and $1$ good performance. Moreover, in order to make each criterion lying in the same interval scale, we normalized them such that $\sum_{k=1}^{|\mathcal{A}^-|} c_i^k = 1$.

\paragraph{Local dataset size (\textbf{base DS})} The first  criterion we considered is the one already used by FedAvg~\cite{mcmahan2017communication} namely the local dataset size given by $c_1^k=|\mathcal{D}_k|/|\cup_{i \in \mathcal{A}^-} \mathcal{D}_i|$. This criterion is a \textit{pure quantitative measure} about the local data, which will serve both as baseline in empirical validation of the results (i.e., when used in isolation) and as part of the entire identified set of criteria in the developed FL system (i.e., when used in a group).

%both as a fundamental scoring criterion and as the baseline in empirical validation of the results.

%Like in FedAvg~\cite{mcmahan2017communication}, we take into account the local dataset size. So, we define $c_1^k=|\mathcal{D}_k|/|\cup_{i \in A} \mathcal{D}_i|$.
\paragraph{Local label diversity (\textbf{Ld})} The second considered criterion is the \textit{diversity of labels} in each local dataset, measuring the diversity of each local dataset in terms of class labels. We assert this criterion to be important since it can provide a clue on how much each device can be useful for learning to predict different labels. To quantify this criterion we use $c_2^k=\delta(\mathcal{D}_k)/\sum_{i \in \mathcal{A}^-}\delta(\mathcal{D}_i)$ where $\delta$ measures the number of different labels (classes) present over the samples of that dataset.

%For this criterion, we consider the normalized version of the ratio $\delta(\mathcal{D}_k)/\delta(\cup_{i \in \mathcal{A}} \mathcal{D}_i)$, where the function $\delta$ measures the number of different labels (classes) present over the samples of that dataset. It follows then, $c_2^k=\delta(\mathcal{D}_k)/\sum_{i \in \mathcal{A}^-}\delta(\mathcal{D}_i)$.

%the number of labels contained in each local dataset, in order to emphasize its \textit{diversity}.

%\paragraph{Training loss}\antonio{to be described, probably we will not use it}

\paragraph{Local model divergence (\textbf{Md})} With non-IID distributions --- and this is the case of our dataset --- model performance dramaticaly gets worse~\cite{noniid}. Moreover, a large number of local training epochs may lead each device to move further away from the initial global model, towards the opposite of the global objective~\cite{sahu2018convergence}. Therefore, a possible solution inspired by~\cite{sahu2018convergence} is to limit these negative effects, by penalizing higher divergences and highlighting local models that are not very far from the received global model. We evaluate the local model divergence as $c_3^k = \varphi^k / \sum_{i \in \mathcal{A}^-}\varphi^i$ where $\varphi^i = \frac{1}{\sqrt{||\mathbf{w}^G-\mathbf{w}^i||_2+1}}$.

%(i.e. a network with two 5x5 convolution layers --- the first with 32 channels, the second with 64, each followed with 2x2 max pooling ---, a fully connected layer with 512 units and ReLu activation, and a final softmax output layer, for a total of 1,663,370 parameters).

%\antonio{I'm working on this part} (i) partition type \yashar{Do we need (i)?}, (ii) aggregation operator and their parameters. In the following, we describe each of these dimensions.

%\subsubsection{Partition type}
%We experiment the behaviour of FL with our weighting protocol both with IID and non-IID partitioned datasets. In the former, we build the local datasets by uniformly sampling the original dataset, but assigning each agent a random number of data. In the latter, we divide the whole dataset in 300 shards of 200 elements of the same class and we assign each agent samples of at most three among the possible classes in order to build a \textit{pathological non-IID distribution}~\cite{mcmahan2017communication}; finally, each local dataset is randomly subsampled in order to artificially simulate the different local dataset sizes.

\section{Results and Discussion}
\label{subsec:eval_FL_proposed}

%\yashar{XXX This section is largely stable and reads nice - one more check is needed, please try to improve the discussion where needed. XXX}\\
In order to validate the empirical performance of the proposed FL system, an extensive set of experiments has been carried out with respect to three under-study exploration dimensions in agreement with the assumption presented in Section  \ref{sec:intro}.
%\begin{itemize}
%    \item  The effect of \textit{individual local criteria} (cf. Sec \ref{ref:study_a})
%    \item The impact of \textit{priority order} in multi-criteria aggregation (cf. Sec \ref{ref:study_b})
%    \item The impact of\textit{ online adjustment} of the priority order in multi-criteria aggregation (cf. Sec \ref{ref:study_c}).
%\end{itemize}
The final results are presented in  Table~\ref{tab:FL_results}. Note that the results are presented for reaching two distinctive desired target global accuracy of 75\% and 80\%.\footnote{We chose these accuracy values  since they represent reasonable accuracy values and prediction tasks higher than 80\% are not reached in the 1,000 allowed rounds of communication.} Each column indicates the percentage of devices which participate to the federation process that reach a desired target accuracy\footnote{The total number of participating devices in the federation is 371, thus 20\%, as an example, indicates the round of communication required for 0.2$\times$317=75 devices to reach the desired target accuracy.}.
%\antonio{Since the total number of participating devices is 371, the cells corresponding to the 20\% column, for instance, indicate the first round of communication in which 75 devices at the same time performed the target accuracy.} 
In addition, we present the results in three groups of (\textbf{Low}, \textbf{Mid}, \textbf{High}) for percentage of participating devices. 
%This was done after noticing correlations in the results of respective groups. 

%In the following, for the simplicity of discussion, we refer to three criteria with baseline dataset size (\textbf{base DS}), model divergence (\textbf{Md}) and label diversity (\textbf{Ld}). 

%Considering two target global accuracies of 75\% and 80\%, Table~\ref{tab:FL_results} provides results of each of these exploration types according to the evaluation approach we defined in Section~\ref{exp_appr}. The results are presented for reaching the desired target accuracies of 75\% (first table) and 80\% (second table). We provide the number of rounds of communication that the experiment indicated in the row needs to allow the percentage of devices in the respective column reach the target accuracy. It can be noticed that we divided columns in three groups (low, mid, high) because of the correlations of results we noticed in considering some specific size ranges for device subsets.

\noindent\textbf{Study A: Effect of individual criteria.} 
\label{ref:study_a}
Study A contemplates answering the question: \dquotes{\textit{Are we able to introduce a set of device- and data- dependent criteria through the help of which we can train a better global model?}}.
 The results for this study are summarized in the row \textbf{Ind} of Table \ref{tab:FL_results}. To answer this question, we considered the effect of each three identified criteria \textbf{base Ds}, \textbf{Md}, \textbf{Ld} \textit{in isolation}. The results with respect to both desired accuracies of 75\% and 80\% show that the new identified criteria (\textbf{Md} and \textbf{Ld}) have an impact in the final quality of the global model, which is \textit{comparable} (in \textbf{Low} and \textbf{Mid} cases) or \textit{superior} with respect to the conventional \textbf{base Ds} criteria (in the case of \textbf{High)}. For example, when comparing \textbf{Md} and \textbf{Ld}, one can notice the results are equal to 25.5 v.s. 27 with a marginal difference of only 6\%. This is while, if we desire to satisfy a higher number of devices (\textbf{High} case) to reach a certain accuracy, the introduced/proposed criteria show a quality substantially better than the \textbf{base Ds} criteria. For example, \textbf{Ld} has a mean performance of 405 compared with 552.5 obtained \textbf{base Ds}. This is equal to an improvement of 36\% with respect to existing baseline.
These initial results already show how the global model can benefit from considering other criteria than just the dataset size. 
%This behavior \yashar{In fact, these superb results, which we obtained on individual evaluation of criteria in the FL setup,} has been the main driving force in our study \yashar{to consider the potential benefits of aggregating of the whole set of criteria (see Study B and Study C)} and \antonio{led us to consider the aggregation of the whole set of criteria, as we experiment in the following}.

%The first exploration considers the aggregation of local models by individually taking into account the properties, in order to test the meaningfulness and the effectiveness of the identified local criteria. Actually, each of the considered criteria exhibits a distinctive behavior, showing that dataset size property faster reaches the target accuracy in a smaller subset of devices; instead, model divergence and, above all, label diversity are able to faster grant the target accuracy for a bigger subset of agents. This behavior has been the boost for searching improvements by gathering the whole set of criteria.

\noindent\textbf{Study B: Impact of Priority order in multi-criteria aggregation.} 
\label{ref:study_b}
Study B focuses on the question: \dquotes{\textit{Are we able to exploit the potential benefits of a prioritized  multi-criteria aggregation operator to  build a more informative global model based on the identified criteria?}}.
The results for this study are summarized in row
\textbf{MCA}
of Table 1. To answer this research question, we performed one experiment for each individual permutation of criteria in the prioritized multi-criteria aggregation setting. Since there are 3 identified criteria, we have in total 6 permutations of criteria. For a fine-grained analysis, we provide the results obtained for \textit{all the permutation runs}, denoted, e.g., by \textbf{Ds} $\succ$ \textbf{Ld} $\succ$ \textbf{Md}, \textbf{Ds} $\succ$ \textbf{Md} $\succ$ \textbf{Ld}. By looking at the results, we can notice that in \textbf{Low} and \textbf{Mid} categories, the best results are obtained for \textbf{Ds} $\succ$ \textbf{Ld} $\succ$ \textbf{Md} and \textbf{Ds} $\succ$ \textbf{Md} $\succ$ \textbf{Ld}. These results share a similar characteristic, which involves the fact that  by considering  \textbf{Ds} as the first important criterion, we can grant a smaller subset of devices the chance to reach to a desired target accuracy in faster pace/rate. This result is in agreement with individual results (see \textbf{Ind} in Table~\ref{tab:FL_results}) in the sense that the \textit{criterion \textbf{Ds} provides the best quality in \textbf{Low} and \textbf{Mid} study cases for both desired target accuracy of 75\% and 80\%}.
%\yashar{tried to be honest but at the same time informative, please check... } 
However, when concentrating on the \textbf{High} category, one can notice \textbf{Md} $\succ$ \textbf{Ds} $\succ$ \textbf{Ld} provides the best performance. This result is a bit surprising and shows that to satisfy a higher number of devices, the criterion \textbf{Md} plays the most important role. This result is surprising from the sense that in the individual results (see \textbf{Ind} in Table~\ref{tab:FL_results}), \textbf{Ld} has the most important performance, while in the obtained result it has the lowest priority. 
%This result can imply the fact \textit{in order to satisfy a higher number of devices to reach a desired accuracy, intuitions/results obtained from individual study may not hold/help for improving the results in the MCA case}. \yashar{Maybe we can improve the last sentence wording.} 
Interestingly, we may notice that in all these best cases, the pattern  \textbf{Ds} $\succ$ \textbf{Ld} always occurs\footnote{We remember here that a preference relation $\succ$ is transitive. Hence \textbf{Ds} $\succ$ \textbf{Md} $\succ$ \textbf{Ld} implies \textbf{Ds} $\succ$ \textbf{Ld}.}.

\noindent\textbf{Study C: Impact of Online Adjustment of the Priority-Order in multi-criteria aggregation.} 
\label{ref:study_c}
Finally, study C studies the question: \dquotes{\textit{Is it possible to update parameters for the aggregation operator (the priority order
of the above-mentioned criteria) via an
online monitoring and adjustment or improving the quality of global model?}}.
The  results  for  this  study  are  summarized  in  row
\textbf{Final}
of  Table ~\ref{tab:FL_results}. This study in fact is concerned with the \textit{dynamic} behavior of our proposed FL approach, by letting the server choose at each round of communication the priority ordering maximizing the accuracy (i.e, obtain the best sub-optimal accuracy). Similar to the previous study, here we also run six experiments, related to the six possible \textit{initializations} for the priority combinations. In Table~\ref{tab:FL_results} we show results related to the best run and to their mean. In this final experimental setting, we see an overall improvement in the performances of the proposes approach when we initialize the priority ordering with \textbf{Md} $\succ$ \textbf{Ds} $\succ$ \textbf{Ld}. Also in this case, the pattern  \textbf{Ds} $\succ$ \textbf{Ld} occurs.

\begin{table}[t!]
\caption{The final results of the empirical evaluation. Each table cell provides the number of rounds of communication necessary to make the percentage of devices (as specified in the columns) reach a desired target accuracy (either 75\% or 80\% in our case). Runs that did not reach the target accuracy for the specified percentage of devices in the allowed rounds (1,000) are marked with \textit{---}. The best results obtained in study \textbf{MCA} are shown in bold {\color{violet} \textbf{violet}} while the best results in study Final, are shown in bold italic {\color{blue} \textbf{\textit{blue}}}.  
%\textit{Abbreviations}: \textbf{Ind} = Individual Criteria, \textbf{MCA} = multi-criteria aggregation, \textbf{Final} =  Final system composed by online adjustment of multi-criteria aggregation priority.
}
\label{tab:FL_results}
\begin{tabular*}{\textwidth}{l@{\extracolsep{\fill}}l@{\extracolsep{\fill}}l@{\extracolsep{\fill}}l@{\extracolsep{\fill}}l@{\extracolsep{\fill}}l@{\extracolsep{\fill}}l@{\extracolsep{\fill}}l@{\extracolsep{\fill}}l@{\extracolsep{\fill}}l@{\extracolsep{\fill}}l}
\toprule
\multicolumn{11}{c}{\textbf{Target accuracy 75\%}} \\
\toprule
&  & \multicolumn{3}{l}{\textbf{Low}} & \multicolumn{3}{l}{\textbf{Mid}} & \multicolumn{3}{l}{\textbf{High}} \\
  \cline{3-5} \cline{6-8} \cline{9-11}
\multicolumn{2}{l}{\textbf{Study/\% devices} }  & 20\% & 30\% & mean & 40\% & 50\% & mean & 70\% & 75\% & mean \\ \hline
\multirow{3}{*}{\rotatebox{0}{\textbf{Ind}}} & \textit{Dataset size (base)} & \textit{22} & \textit{29} & \textit{25.5} & \textit{39} & \textit{62} & \textit{50.5} & \textit{304} & \textit{801} & \textit{552.5} \\ %\cline{2-11} 
 & Model divergence & 24 & 30 & 27 & 41 & 67 & 54 & 274 & 768 & 521 \\ %\cline{2-11} 
 & Label diversity & 25 & 32 & 28.5 & 43 & 70 & 56.5 & 278 & 532 & 405 \\ \hline
\multirow{6}{*}{\rotatebox{0}{\textbf{MCA}}} & Ds $\succ$ Ld $\succ$ Md & \textbf{\color{violet} 20} & \textbf{\color{violet} 29} & \textbf{\color{violet} 24.5} & \textbf{\color{violet} 39} & \textbf{\color{violet} 60} & \textbf{\color{violet} 49.5} & 300 & 823 & 561.5 \\ %\cline{2-11} 
 & Ds $\succ$ Md $\succ$ Ld & \textbf{\color{violet} 20} & \textbf{\color{violet} 29} & \textbf{\color{violet} 24.5} & \textbf{\color{violet} 39} & \textbf{\color{violet} 60} & \textbf{\color{violet} 49.5} & 300 & 669 & 484.5 \\ %\cline{2-11} 
 & Ld $\succ$ Ds $\succ$ Md & 24 & 31 & 27.5 & 41 & 68 & 54.5 & 259 & 768 & 513.5 \\ %\cline{2-11} 
 & Md $\succ$ Ds $\succ$ Ld & 24 & 32 & 28 & 45 & 70 & 57.5 & \textbf{\color{violet} 255} & \textbf{\color{violet} 532} & \textbf{\color{violet} 393.5} \\ %\cline{2-11} 
 & Ld $\succ$ Md $\succ$ Ds & 23 & 30 & 26.5 & 41 & 68 & 54.5 & 270 & 729 & 499.5 \\ %\cline{2-11} 
 & Md $\succ$ Ld $\succ$ Ds & 24 & 32 & 28 & 46 & 70 & 58 & 255 & 620 & 437.5 \\ \cline{2-11}
 & mean & 22.5 & 30.5 & 26.5 & 41.8 & 66 & 53.9 & 273.17 & 690.1 & 481.6 \\ \hline
\multirow{2}{*}{\rotatebox{0}{\textbf{Final}}} & %Ds $\succ$ Ld $\succ$ Md & 21 & 27 & 24 & 39 & 64 & 51.5 & 208 & 552 & 380 \\ \cline{2-11} 
% & Ds $\succ$ Md $\succ$ Ld & 21 & 29 & 25 & 40 & 64 & 52 & 221 & 470 & 345.5 \\ \cline{2-11} 
% & Ld $\succ$ Ds $\succ$ Md & 21 & 28 & 24.5 & 41 & 63 & 52 & 208 & 543 & 375.5 \\ \cline{2-11} 
 Md $\succ$ Ds $\succ$ Ld & \textbf{\color{blue} \textit{12}} & \textbf{\color{blue} \textit{19}} & \textbf{\color{blue} \textit{15.5}} & \textbf{\color{blue} \textit{26}} & \textbf{\color{blue} \textit{57}} & \textbf{\color{blue} \textit{41.5}} & \textbf{\color{blue} \textit{164}} & \textbf{\color{blue} \textit{494}} & \textbf{\color{blue} \textit{329}} \\ \cline{2-11}
% & Ld $\succ$ Md $\succ$ Ds & 24 & 31 & 27.5 & 43 & 62 & 52.5 & 274 & 823 & 548.5 \\ \cline{2-11} 
% & Md $\succ$ Ld $\succ$ Ds & 24 & 31 & 27.5 & 43 & 61 & 52 & 263 & 789 & 526 \\ \hline
 & mean & 20.5 & 27.5 & 24 & 38.6 & 61.8 & 50.2 & 223 & 611.8 & 417.4 \\
 \bottomrule
 \end{tabular*}

\begin{tabular*}{\textwidth}{l@{\extracolsep{\fill}}l@{\extracolsep{\fill}}l@{\extracolsep{\fill}}l@{\extracolsep{\fill}}l@{\extracolsep{\fill}}l@{\extracolsep{\fill}}l@{\extracolsep{\fill}}l@{\extracolsep{\fill}}l@{\extracolsep{\fill}}l@{\extracolsep{\fill}}l}
\toprule
\multicolumn{11}{c}{\textbf{Target accuracy 80\%}} \\
\toprule
&  & \multicolumn{3}{l}{\textbf{Low}} & \multicolumn{3}{l}{\textbf{Mid}} & \multicolumn{3}{l}{\textbf{High}} \\
 \cline{3-5} \cline{6-8} \cline{9-11}
\multicolumn{2}{l}{\textbf{Study/\% devices} }  & 20\% & 30\% & mean & 40\% & 50\% & mean & 70\% & 75\% & mean \\ \hline
\multirow{3}{*}{\textbf{Ind}} & \textit{Dataset size (base)} & \textit{31} & \textit{45} & \textit{38} & \textit{72} & \textit{136} & \textit{104} & \textit{---} & \textit{---} & \textit{---} \\
 & Model divergence & 31 & 46 & 38.5 & 82 & 151 & 116.5 & --- & --- & --- \\ 
 & Label diversity & 36 & 53 & 44.5 & 90 & 161 & 125.5 & --- & --- & --- \\ \hline
\multirow{6}{*}{\textbf{MCA}} & Ds $\succ$ Ld $\succ$ Md & \textbf{\color{violet} 30} & \textbf{\color{violet} 45} & \textbf{\color{violet} 37.5} & \textbf{\color{violet} 72} & \textbf{\color{violet} 135} & \textbf{\color{violet} 103.5} & --- & --- & --- \\ 
 & Ds $\succ$ Md $\succ$ Ld & \textbf{\color{violet} 30} & \textbf{\color{violet} 45} & \textbf{\color{violet} 37.5} & \textbf{\color{violet} 72} & \textbf{\color{violet} 135} & \textbf{\color{violet} 103.5} & --- & --- & --- \\ 
 & Ld $\succ$ Ds $\succ$ Md & 31 & 46 & 38.5 & 82 & 149 & 115.5 & --- & --- & --- \\ 
 & Md $\succ$ Ds $\succ$ Ld & 36 & 53 & 44.5 & 84 & 161 & 122.5 & --- & --- & --- \\  
 & Ld $\succ$ Md $\succ$ Ds & 31 & 46 & 38.5 & 82 & 151 & 116.5 & --- & --- & --- \\  
 & Md $\succ$ Ld $\succ$ Ds & 36 & 53 & 44.5 & 90 & 161 & 125.5 & --- & --- & --- \\ \cline{2-11}
 & mean & 32.3 & 48 & 40.1 & 80.3 & 148.6 & 114.5 & --- & --- & --- \\ \hline
\multirow{2}{*}{\rotatebox{0}{\textbf{Final}}}%& Ds $\succ$ Ld $\succ$ Md & 31 & 44 & 37.5 & 81 & 146 & 113.5 & - & - & - \\ \cline{2-11} 
% & Ds $\succ$ Md $\succ$ Ld & 32 & 44 & 38 & 82 & 140 & 111 & - & - & - \\ \cline{2-11} 
% & Ld $\succ$ Ds $\succ$ Md & 32 & 44 & 38 & 83 & 152 & 117.5 & - & - & - \\ \cline{2-11} 
 & Md $\succ$ Ds $\succ$ Ld & \textbf{\color{blue} \textit{21}} & \textbf{\color{blue} \textit{36}} & \textbf{\color{blue} \textit{28.5}}& \textbf{\color{blue} \textit{61} }& \textbf{\color{blue} \textit{133}} & \textbf{\color{blue} \textit{97}} & --- & --- & --- \\ \cline{2-11} 
% & Ld $\succ$ Md $\succ$ Ds & 32 & 46 & 39 & 82 & 151 & 116.5 & - & - & - \\ \cline{2-11} 
% & Md $\succ$ Ld $\succ$ Ds & 32 & 47 & 39.5 & 80 & 134 & 107 & - & - & - \\ \hdashline
 & mean & 30 & 43.5 & 36.7 & 78.1 & 142.6 & 110.4 & --- & --- & --- \\ \bottomrule
\end{tabular*}
\end{table}

\section{Conclusions and Future perspectives}
\label{sec:conlusion}
%\yashar{I have drafted this, please check and improve}\\
In this work, we presented a practical protocol for effectively aggregating data by proposing a set of \textit{device-} and \textit{data-aware} properties (criteria) that are exploited by a central server in order to obtain a more qualitative/informative global model. %\yashar{we can even drop the following} The main contributions of the proposed protocol include, first: we proposed a formal protocol to aggregated local data by relying on set of device-dependent criteria, second: \yashar{mention criteria}, third: \yashar{mention aggregation}, finally (\yashar{mention online adj}).
Our experiments show that the standard federated learning standard, FedAvg can be substantially improved by training high-quality models using relatively few rounds of communication, by using a properly defined set of local criteria and using aggregation strategy that can exploit the information from such criteria.
%\yashar{till here}
%\antonio{This is my proposal}
%Over the last few years, Federated Learning has been proposed as a mean to improve privacy for distributed data/users in machine learning scenarios. In this work, we incrementally proposed some novelties for Federated Learning, aiming to incorporate some domain- and client-specific knowledge for giving the domain expertthe possibility to leverage that information for improving the model with respect to specific objectives. More specifically, we provided the domain expert with the possibility to set the contribution of each device to the federated model based on their adherence to some target criteria, instead of using just the local dataset size. Moreover, we proposed a formal protocol to compute such contribution by means of a priority-aware aggregation operator which, for each device, aggregates local criteria evaluations; finally, in order to maximize the performances, we also evaluated the possibility to automatically adjust the parameters of that operator via a local search strategy.
%Results showed that global model is dramaticaly sensitive to use different information than the dataset size in merging the computed local models. Moreover, both the multi-criteria aggregation for computing device contribution and the online adjustment of aggregation parameters showed superb results with respect to the objective we set by means of our identified criteria
%\yashar{the following are suggestion by anotnio}
Future perspectives for this work concern with the identification of other local criteria --- both general purpose and domain-specific ---, the experimentation with other aggregation operators and with other interesting datasets, as well as the extension of this federated approach to other machine learning systems, such as those in recommendation domain.

\paragraph{Acknowledgements} The authors wish to thank Angelo Schiavone for fruitful discussions and for helping with the implementation of the framework.
%
% The next two lines define the bibliography style to be used, and the bibliography file.
\bibliographystyle{splncs04}
\bibliography{base}

\end{document}